# Optimization of Convolutional Neural Network Using the Linearly Decreasing Weight Particle Swarm Optimization

Tatsuki Serizawa[*1*2] Hamido Fujita[*1]
[*1] *Iwate Prefectural University* [*2] Welmo Inc.

**Abstract.** Convolutional neural network (CNN) is one of the most frequently used deep learning techniques. Various forms of models have been proposed and improved for learning at CNN. When learning with CNN, it is necessary to determine the optimal hyperparameters. However, the number of hyperparameters is so large that it is difficult to do it manually, so much research has been done on automation. A method that uses metaheuristic algorithms is attracting attention in research on hyperparameter optimization. Metaheuristic algorithms are naturally inspired and include evolution strategies, genetic algorithms, antcolony optimization and particle swarm optimization. In particular, particle swarm optimization converges faster than genetic algorithms, and various models have been proposed. In this paper, we propose CNN hyperparameter optimization with linearly decreasing weight particle swarm optimization (LDWPSO). In the experiment, the MNIST data set and CIFAR-10 data set, which are often used as benchmark data sets, are used. By optimizing CNN hyperparameters with LDWPSO, learning the MNIST and CIFAR-10 datasets, we compare the accuracy with a standard CNN based on LeNet-5. As a result, when using the MNIST dataset, the baseline CNN is 94.02% at the 5th epoch, compared to 98.95% for LDWPSO CNN, which improves accuracy. When using the CIFAR-10 dataset, the Baseline CNN is 28.07% at the 10th epoch, compared to 69.37% for the LDWPSO CNN, which greatly improves accuracy. This paper is presented at the 36th Annual Conference of the Japanese Society for Artificial Intelligence. The final version is available at the following URL: https://doi.org/10.11517/pjsai.JSAI2022.0_2S4IS2b03

## 1. Introduction

Deep learning is one of the research fields that is being worked on in the field of artificial intelligence. Deep learning can be categorized primarily into discriminative models, generative models, and hybrid models (L. Deng and D. Yu 2013). For example, the discriminant model is deep neural network (DNN), convolutional neural network (CNN), or recurrent neural network (RNN). Generating models include deep belief networks (DBN), restricted Boltzmann machines, and so on. Also, a hybrid model is a structure that combines an identification model and a generation model. In this, various forms of the analysis process using machine learning have been proposed and improved. As an example, (M. Mohamad et al. 2020) propose an analysis process based on correlation-

Contact: Tatsuki Serizawa, Welmo Inc.
E-mail: TSerizawa577@gmail.com

based feature selection, a best-first search algorithm, a soft set, and two integrated models combined with mutually complementary rough-set theories. However, when learning with machine learning methods such as CNN, it is necessary to determine the optimal hyperparameters. There are so many hyperparameters in deep learning, and it is difficult to determine the optimum value manually, so much research has been done on automation.

There is a metaheuristic algorithm as an automation method. Metaheuristic algorithms are a way to solve difficult optimization problems, and in recent years have also been used to optimize CNNs (R. Oullette et al. 2004; L. M. R. Rere et al. 2015; L. M. R. Rere et al. 2016; G. Rosa et.al. 2015; Y. Zhining et al. 2015; Vina Ayumi et al. 2016). This algorithm is inspired by nature and is based on phenomena in physics, biology, animal behavior and so on. Algorithms based on biological phenomena include evolutionary strategies (ES) and genetic algorithms (GA). Based on physics phenomena, there are microcanonical annealing (MA) and simulated annealing (SA). Based on animal behavior, there are Ant Colony Optimization (ACO), Firefly Algorithm (FA), Bat Algorithm (BA) and Particle Swarm Optimization (PSO).

Among them, various forms of PSO have been proposed for network optimization. PSO converges faster than GA and is attracting attention as an excellent method (A. Sahu et al. 2012; Wang Hu et al. 2015). In the study of (da Silva GLF et al. 2018), they optimized the hyperparameter of CNN used to classify pulmonary nodule candidate images into nodule and non-nodule by using simple PSO. In the study of (Wei-Chang Yeh et al. 2013), the weight of the artificial neural networks model has been optimized using a new method called improved Parameter-Free Simplified Swarm Optimization (SSO), and excellent performance has been obtained. Also, in PSO, the weight parameter is an important parameter for optimization. Among them, the linear decreasing inertia weight proposed in (Y.H. Shi et al. 1999; Y.H. Shi et al. 2000; R.C. Eberhart et al. 2000) is an excellent method for converging PSO efficiently and is used in many research. In this paper, CNN hyperparameter optimization is performed by using linearly decreasing weight PSO (LDWPSO). By performing hyperparameter optimization using LDWPSO, CNN parameters can be efficiently determined. We aim to improve the accuracy of the MNIST and CIFAR-10 data sets, which are often used as benchmarks, by conducting experiments using CNN based on LeNet-5.

The outline of this paper is as follows; Section 2 describes PSO, CNN and related research. Section 3 describes the proposed method and Section 4 describes the experiment and its results. The final section 5 describes concludes this paper.

## 2. Related Works

### *2.1. Particle Swarm Optimization*

PSO is one of the heuristic algorithms proposed by James Kennedy and Russell Eberhart (1995). The PSO algorithm is shown in Algorithm 1.

| | **Algorithm 1**: Particle Swarm Optimization |
|---|---|
| 1 | iter = 0 |
| 2 | Initialize v and x of all particles |
| 3 | Initialize pBest and gBest |
| 4 | |
| 5 | while iter $\leq$ $iter_{max}$ do |

6 for i = 1 to N do
7 for j = 1 to D do
8 Update the velocity and position of the particles from Eq. 1 and 2.
9 end for
10 Calculate evaluation value of particle i
11 if $f\left(x_i^{iter+1}\right) < f(pBest_i^{iter})$ then
12 $pBest_i^{iter+1} = x_i^{iter+1}$
13 end if
14 end for
15 $k = \arg min\, f(pBest_i^{iter+1}$
16 if $f(pBest_k^{iter+1}) < f(gBest^{iter})$ then
17 $gBest^{iter+1} = pBest_k^{iter+1}$
18 end if
19 t = t + 1
20 end while

Where $iter$ is the number of iterations and N is the total number of particles. Each particle has its current position and velocity, and the best position they find is called the personal best (pBest), and the best particle position in the whole group is called the global best (gBest). Each particle moves so that its new position approaches gBest and pBest, thereby searching for the optimal solution. The velocity v of each particle in movement is updated by Eq. (1).

$$v_{id}(t+1) = w \times v_{id}(t) + c_p \times r_p \times \left(pBest_{id} - x_{id}(t)\right) + c_g \times r_g \times \left(gBest_{id} - x_{id}(t)\right) \quad (1)$$

Where $v_{id}$ is the velocity of the i-th particle in the d dimension, x is the current particle position, w is the weight, $c_p$ and $c_g$ are predefined constants called acceleration factors, and $r_p$ and $r_g$ are random number [0 : 1]. By calculating $v(t+1)$ using Eq. (1), the position is updated as shown in Eq. (2).

$$x_{id}(t+1) = x_{id}(t) + v_{id}(t+1) \quad (2)$$

It has been found that the weight parameter w used in Eq. (1) has a significant effect on convergence, and various equations have been proposed. The weight parameter has been originally introduced by (Y.H. Shi et al. 1998). They found that use a constant w-value range, and at w < 0.8, PSO tends to trap at a local optimum. (R.C. Eberhart et al. 2001) uses random values for weights to enable PSO to track optimal values in a dynamic environment. (Y. Feng, G Teng et al. 2007; Y. Feng, Y.M. Yao et al. 2007) uses a weighted chaotic model in which the chaotic term is added to a linearly decreasing weight. (Y.H. Shi et al. 1999; Y.H. Shi et al. 2000; R.C. Eberhart et al. 2000) uses a strategy where weight values are determined based on the number of iterations. This is expressed as Eq. (3).

$$\mathrm{w(iter)} = \frac{iter_{max} - iter}{iter_{max}}(w_{max} - w_{min}) + w_{min} \quad (3)$$

$iter_{max}$ is the maximum number of iterations the PSO is allowed to continue. $w_{max}$ and $w_{min}$ are the maximum and minimum weights to be specified, and 0.9 and 0.4 are generally specified. This linearly decreasing weight PSO is used in many optimization tasks and is a common technique. In this paper, optimization is performed by LDWPSO.

### *2.2. Convolutional Neural Network*

CNN is a type of feed-forward DNN, especially used in image recognition tasks. Several models have been proposed for the CNN architecture, such as ResNet proposed by (Kaiming He et al. 2015), DenseNet (Gao Huang, 2017), and VGG (Karen Simonyan, 2015). In this study, as shown in Fig. 1, we will construct an architecture based on LeNet-5 of (Y. LeCun, 1998), which is one of the most basic architectures. It consists of two convolution layers, two pooling layers, two fully connected layers and an output layer.

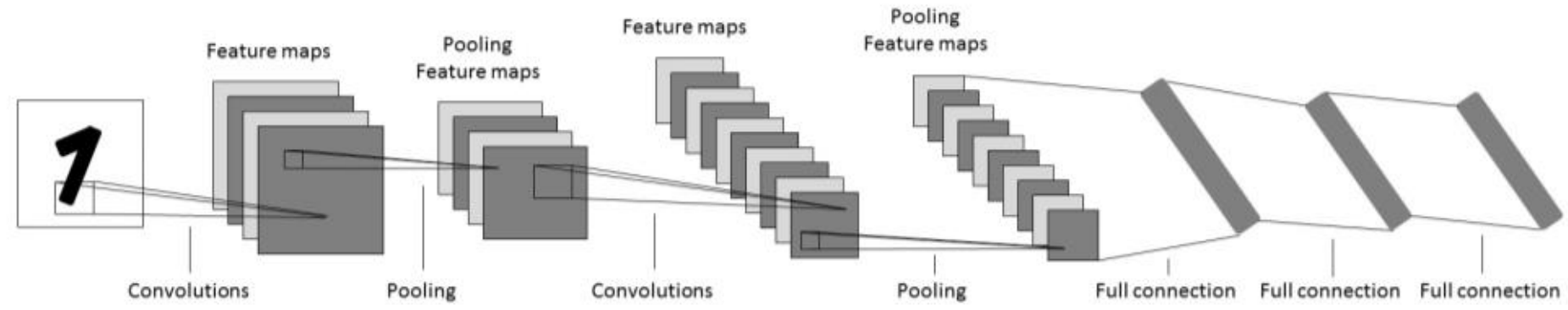

**Fig. 1.** CNN structure based on LeNet-5

The convolution layer performs convolution operations. A convolution operation is performed by applying a filter to input data. The product-sum operation is performed while sliding the filter on the input data and stored in the output corresponding location. The result obtained by the convolution operation is called a feature map. The pooling layer is an operation that spatially downsamples the input feature map. For example, if 2× 2 max pooling is performed with stride 2 for a 4×4 feature map, a 2×2 feature map will be generated. The full connected layer receives the feature map generated from the convolution layer and the pooling layer and passes it to the output layer. For classification tasks, the output layer has each neuron assigned to one class label. The MNIST and CIFAR-10 datasets used in this experiment are composed of 10 neurons corresponding to the number of classes.

## 3. Proposed Method

In this paper, we use LDWPSO to optimize CNN parameters and improve the accuracy obtained by CNN. The flow of the proposed method is shown in Fig. 2.

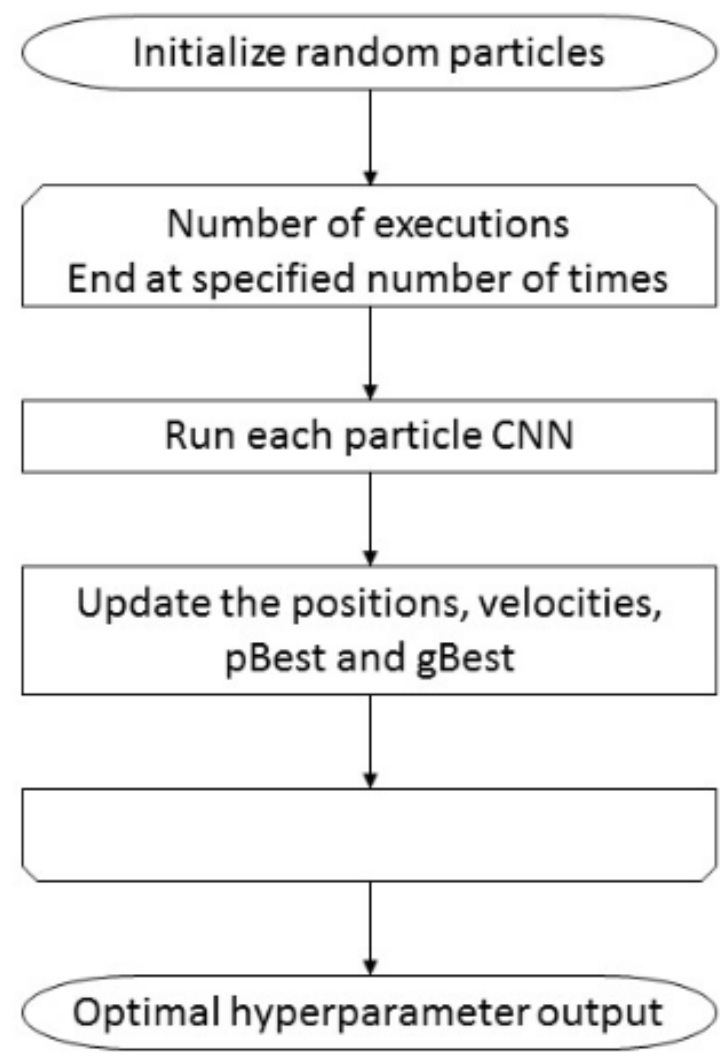

**Fig. 2**: Flow chart of the proposed method

In the experiment, first set the position and velocity of each particle. Next, CNN is executed for each particle, and the position, velocity, pBest, and gBest are updated based on the obtained results. This operation is repeated the specified number of times, and the gBest particle parameter obtained as the last result is output as the optimum parameter. The loss function used in this experiment is cross-entropy and is expressed by the following Eq. (4).

$$\mathrm{E(w)} = -\frac{1}{N}\sum_{n=1}^{N}\sum_{k=1}^{K} d_{nk}\log y_{nk} \tag{4}$$

Where N is the number of training data samples, K is the number of outputs, d is the desired output for training data for x, and y is the network output for input. Also, accuracy is used as the evaluation function.

In the proposed method, the main configuration of the CNN architecture is LeNet-5. This is a relatively simple architecture consisting of 2 convolutional layers and 2 fully connected layers, and use because it is often used as a comparison architecture in experiments. The pooling layer uses max pooling. Table 2 shows the baseline parameters and optimization parameters of LeNet-5 used for comparison when the convolution layers are C1 and C2, respectively, and the total coupling layers are FC1 and FC2. The number of filters in the convolution layers C1 and C2 is optimized between 4 and 100, and the kernel size is 3, 5 and 7. The number of neurons in full connect layers optimizes between 4 and 200. The activation function of each layer uses sigmoid, relu, and tanh, and the batch size is optimized from 10 to 100. The Optimizer uses Adam or Stochastic Gradient Descent (SGD) with a learning rate of 0.01, respectively. In building the model, we used the open source neural network library, Keras.

The main parameters of LDWPSO use for optimization are shown in Table 3. The Swarm size is 10 and the maximum number of iterations is 10. The Cognitive parameter and Social parameter are both 2.0, and the weight decreases linearly from 0.9 to 0.4. It ends when the maximum iteration is reached as an end condition.

**Table 2** Particle definition in our methodology

| Hyperparameter | Baseline | Optimization Value |
|---|---|---|
| Number of filters in C1 | 6 | 4 – 100 |
| Number of filters in C2 | 16 | 4 – 100 |
| Size of kernel in C1 | 5 | 3, 5, 7 |
| Size of kernel in C2 | 5 | 3, 5, 7 |
| Activation function in C1 | sigmoid | sigmoid, relu, tanh |
| Activation function in C2 | sigmoid | sigmoid, relu, tanh |
| Activation function in FC1 | sigmoid | sigmoid, relu, tanh |
| Activation function in FC2 | sigmoid | sigmoid, relu, tanh |
| Number of neurons in FC1 | 120 | 4 – 200 |
| Number of neurons in FC2 | 84 | 4 – 200 |
| Batch size in the training | 10 | 10 – 100 |
| Optimizer | SGD | Adam, SGD |

**Table 3** Parameters for PSO algorithm

| LDWPSO Parameter | Value |
|---|---|
| Swarm size | 10 |
| Number of iterations | 10 |
| Cognitive parameter | 2.0 |
| Social parameter | 2.0 |
| Inertia weight | 0.9 – 0,4 |

All experiments are performed on a machine with an Intel Core i5-6600K processor, 32GB RAM, GeForce GTX TITAN X-12GB.

## 4. Experiment and Results

In this study, we experiment with two datasets. The first experiment use the MNIST dataset, and the second experiment use the CIFAR-10 dataset. Fig. 3 shows an example of each dataset.

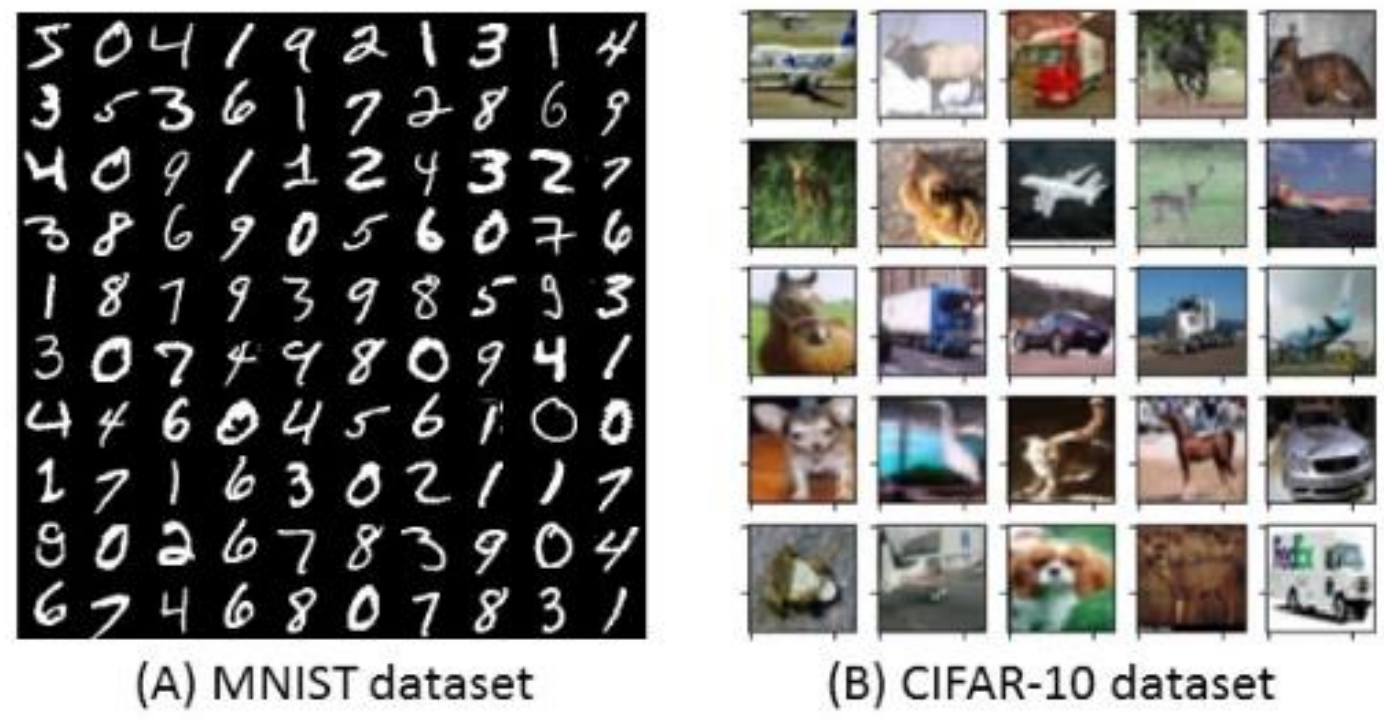

**Fig. 3** Example of some image; (A) MNIST dataset, (B) CIFAR-10 dataset

*4.1. Experiment using MNIST dataset*

MNIST is an image dataset that collects 60000 handwritten number images for learning and 10000 for testing. Each image is given a correct label from 0 to 9. In the experiment, optimization has been performed with LDWPSO every 5 epochs, and learning is performed based on the obtained parameters. Each experiment is performed 30 times, and the average value is taken.

Accuracy and its variance are shown in Table 4 and Fig. 4. When learning 5 epoch, the accuracy of CNN optimized with LDWPSO has 98.95%, which is higher than 94.02% of CNN without optimization. When looking at any epoch, it is found that LDWPSO CNN can obtain higher accuracy than Baseline CNN. Especially in the 1st epoch, Baseline CNN is 18.45%, while LDWPSO CNN has high accuracy of 97.59%, and the variance value is also low at 0.10, so it can see that the optimized CNN converges to high accuracy at an early stage. From this, it is thought that the proposed LDWPSO CNN can find the optimal parameters and can obtain better results compared to the baseline CNN. Also, the average optimization calculation time for LDWPSO CNN is 2381s.

**Table 4** Accuracy and Dispersion of Baseline CNN and LDWPSO CNN using MNIST dataset

| Epoch | Baseline CNN | | LDWPSO CNN | |
|---|---|---|---|---|
| | Accuracy | Dispersion | Accuracy | Dispersion |
| 1 | 18.45 | 45.64 | 97.59 | 0.10 |
| 2 | 75.25 | 36.14 | 98.43 | 0.02 |
| 3 | 89.10 | 2.02 | 98.61 | 0.07 |
| 4 | 92.57 | 0.48 | 98.80 | 0.07 |
| 5 | 94.02 | 0.25 | 98.95 | 0.02 |

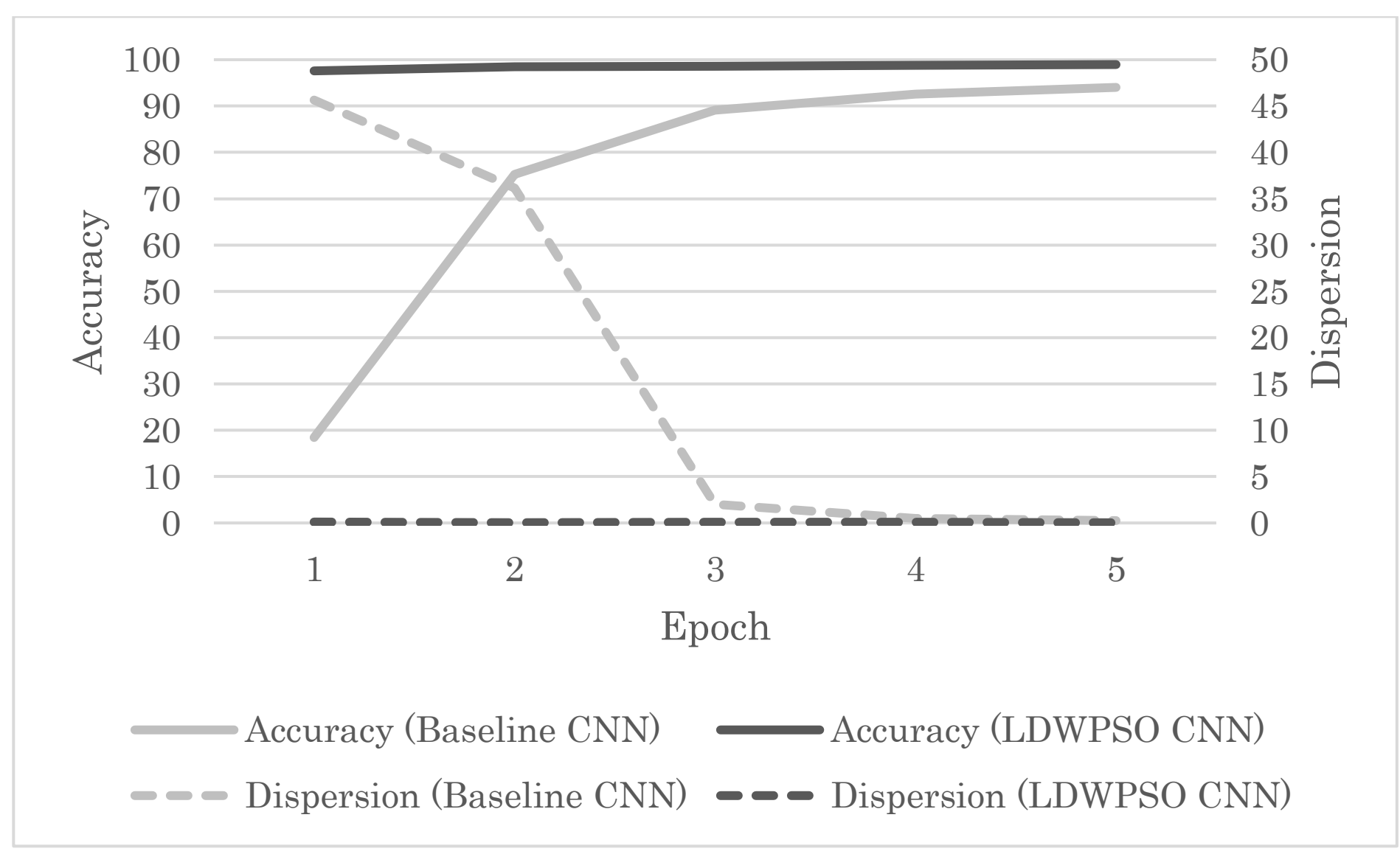

**Fig. 4** Accuracy and dispersion of Baseline CNN and LDWPSO CNN using MNIST dataset

### 4.2. Experiment using CIFAR-10 dataset

The CIFAR-10 dataset is a collection of 50000 images for learning and 10000 images for testing. It is often used as a benchmark for object recognition and is given a 10-class label. In the experiment, optimization is performed with LDWPSO every 10 epochs, and learning is performed based on the obtained parameters. Each experiment is performed 30 times, and the average value is taken.

Accuracy and its variance are shown in Table 5 and Fig. 5. When learning 10 epoch, LDWPSO CNN accuracy is 68.92%, much higher than Baseline CNN 22.21%. When looking at any epoch, both LDWPSO CNNs are more accurate than Baseline CNNs. Also, looking at the variance, Baseline CNN is as high as 24.88 at 5 epoch, and learning is not done expective, while LDWPSO CNN is the highest at 3 epoch and is 1.49, and learning is well done expective. This suggests that the LDWPSO CNN has found parameters that can be successfully learned through optimization and has obtained good results. Also, the average optimization calculation time for LDWPSO CNN is 8547s.

**Table 5** Accuracy and Dispersion of Baseline CNN and LDWPSO CNN using CIFAR-10 dataset

| Epoch | Baseline CNN | | LDWPSO CNN | |
|---|---|---|---|---|
| | Accuracy | Dispersion | Accuracy | Dispersion |
| 1 | 10.02 | 0.01 | 51.80 | 0.74 |
| 2 | 10.11 | 0.21 | 59.59 | 0.94 |
| 3 | 10.28 | 0.64 | 63.58 | 1.49 |
| 4 | 11.31 | 3.72 | 65.94 | 1.14 |
| 5 | 14.50 | 24.88 | 67.51 | 0.81 |
| 6 | 18.91 | 25.21 | 68.50 | 0.48 |
| 7 | 22.21 | 16.51 | 68.92 | 0.72 |
| 8 | 24.54 | 5.79 | 69.18 | 0.75 |
| 9 | 26.45 | 3.12 | 69.53 | 0.41 |
| 10 | 28.07 | 4.43 | 69.37 | 0.41 |

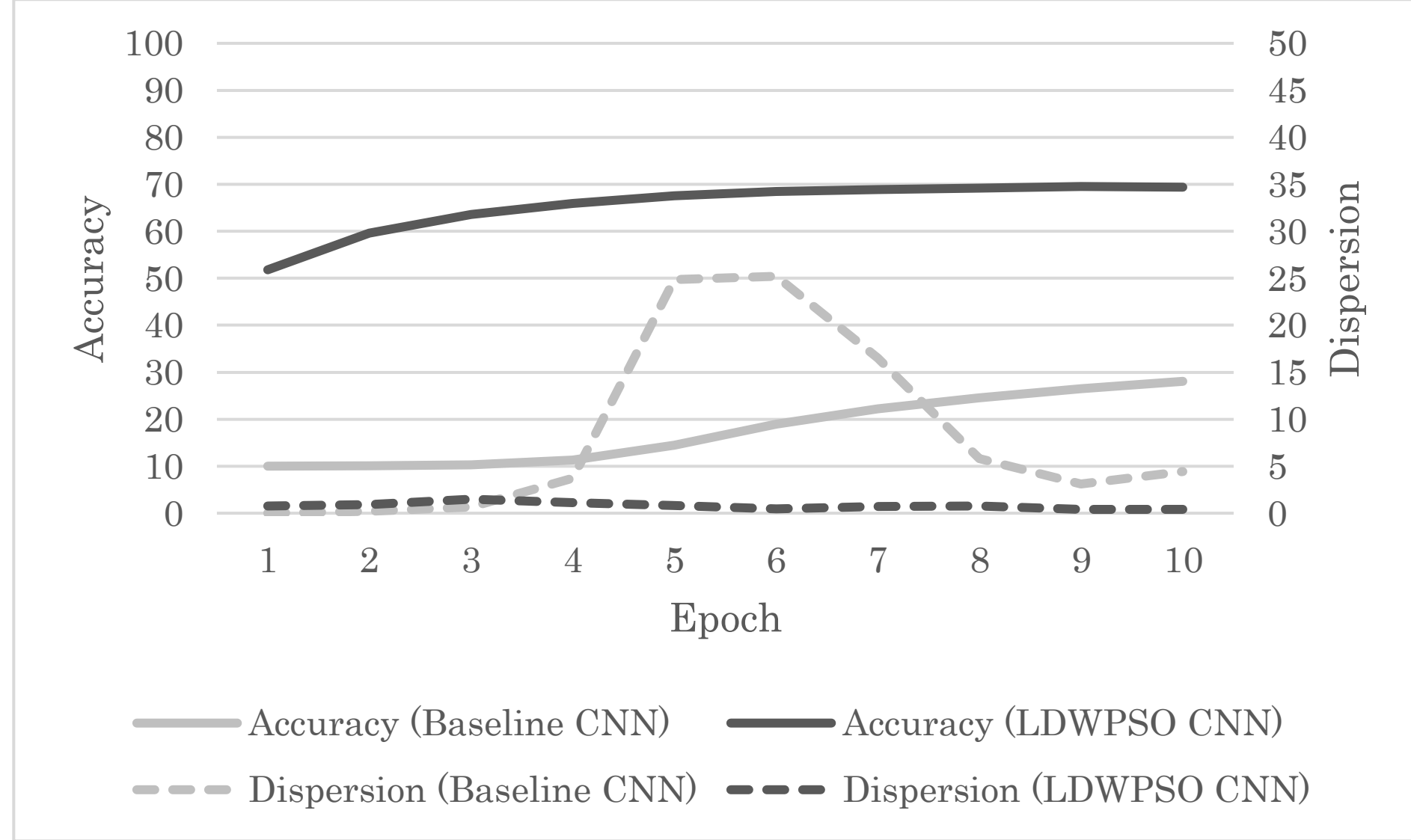

**Fig. 5** Accuracy and dispersion of Baseline CNN and LDWPSO CNN using CIFAR-10 dataset

## 5. Conclusion

In this paper, we propose a method of using linearly decreasing weights for PSO, which is one of the metaheuristic algorithms, for hyperparameter optimization of CNN. Experiments are performed using the MNIST and CIFAR-10 datasets, and both achieved better results than CNN without LDWPSO. In the results using the MNIST dataset, the accuracy of Baseline CNN is 94.02% at the 5th epoch, compared to 98.95% for LDWPSO CNN. Baseline CNN has a higher variance value at the beginning of the first epoch, such as 45.64, while LDWPSO CNN is 0.1 at the highest epoch, and it has been found that it converged with high accuracy from the first epoch. In the results using the CIFAR-10 data set, the accuracy of Baseline CNN is 28.07% at the 10th epoch, while that of LDWPSO CNN is 69.37%, which greatly improved the accuracy. In addition, Baseline CNN has a large variance value of 25.21 at the 6th epoch, and stable learning is not possible. On the other hand, LDWPSO CNN is 1.49 at the 3rd epoch at the highest and can be stably learned. From these results, it is considered that the optimization by LDWPSO can find excellent hyperparameters and can provide high classification accuracy.

As future work, it is necessary to investigate the effectiveness of the proposed method by adapting it to research on the latest image classification. In addition, after adding a PSO termination condition, you should investigate whether the proposed method is valid by comparing simple PSO and LDWPSO. Furthermore, in this research, we conducted experiments on the possibility of optimizing with LDWPSO CNN. Therefore, it is necessary to experiment with other optimization methods and CNN architectures to compare the effectiveness of computational cost and complexity. In addition, it is necessary to investigate whether the accuracy can be demonstrated even in cases other than image datasets by conducting experiments using benchmark data sets other than images.